\documentclass[conference]{IEEEtran}
\IEEEoverridecommandlockouts

\usepackage{cite}
\usepackage{amsmath,amssymb,amsfonts}
\usepackage{algorithmic}
\usepackage{algorithm}
\usepackage{graphicx}
\usepackage{textcomp}
\usepackage{xcolor}
\usepackage{booktabs}
\usepackage{multirow}
\usepackage{graphicx}
\usepackage{subcaption}

\def\BibTeX{{\rm B\kern-.05em{\sc i\kern-.025em b}\kern-.08em
    T\kern-.1667em\lower.7ex\hbox{E}\kern-.125emX}}

\begin{document}

\title{UAV-MARL: Multi-Agent Reinforcement Learning for Time-Critical and Dynamic Medical Supply Delivery}

\author{
\IEEEauthorblockN{Islam Guven, Mehmet Parlak}
  \\
  \IEEEauthorblockA{
    \textit{ICTEAM, Université catholique de Louvain}\\
    Ottignies-Louvain-la-Neuve, Belgium \\
    islam.guven@uclouvain.be
  }
}

\maketitle

\begin{abstract}
Unmanned aerial vehicles (UAVs) are increasingly used to support time-critical medical supply delivery, providing rapid and flexible logistics during emergencies and resource shortages. However, effective deployment of UAV fleets requires coordination mechanisms capable of prioritizing medical requests, allocating limited aerial resources, and adapting delivery schedules under uncertain operational conditions. This paper presents a multi-agent reinforcement learning (MARL) framework for coordinating UAV fleets in stochastic medical delivery scenarios where requests vary in urgency, location, and delivery deadlines. The problem is formulated as a partially observable Markov decision process (POMDP) in which UAV agents maintain awareness of medical delivery demands while having limited visibility of other agents due to communication and localization constraints. The proposed framework employs Proximal Policy Optimization (PPO) as the primary learning algorithm and evaluates several variants, including asynchronous extensions, classical actor--critic methods, and architectural modifications to analyze scalability and performance trade-offs. The model is evaluated using real-world geographic data from selected clinics and hospitals extracted from the OpenStreetMap dataset. The framework provides a decision-support layer that prioritizes medical tasks, reallocates UAV resources in real time, and assists healthcare personnel in managing urgent logistics. Experimental results show that classical PPO achieves superior coordination performance compared to asynchronous and sequential learning strategies, highlighting the potential of reinforcement learning for adaptive and scalable UAV-assisted healthcare logistics.\end{abstract}

\begin{IEEEkeywords}
Multi-agent reinforcement learning (MARL), UAV coordination, swarm, autonomous drone delivery, medical supply delivery, healthcare logistics, dynamic task allocation, proximal policy optimization, reward shaping, time-critical delivery, stochastic logistics, drone delivery systems.
\end{IEEEkeywords}

\section{Introduction}

Unmanned aerial vehicles (UAVs) are increasingly utilized in autonomous navigation, mission-critical data collection, and real-time environmental monitoring applications such as precision agriculture \cite{precision}. Beyond sensing and monitoring tasks, UAVs are also emerging as a promising solution for time-critical logistics, particularly in the distribution of medical supplies from central depots to healthcare facilities. Such operations require the coordination of multiple delivery vehicles under strict time constraints and payload limitations. The challenge becomes even more critical during epidemic outbreaks, natural disasters, or supply chain disruptions, when ground transportation infrastructure may be compromised and rapid response is essential for patient outcomes. Although UAVs provide rapid and flexible delivery capabilities independent of road networks, efficient coordination of multiple UAVs under dynamic operational constraints remains an open research problem, particularly in time-critical healthcare logistics.

Reliable medical supply chains are essential for maintaining effective healthcare services, particularly where rapid and flexible delivery of critical resources is required. Intelligent decision-support systems are therefore needed to assist medical practitioners in allocating limited resources and coordinating logistics operations efficiently \cite{garg2023lastmile,sushma2025spatial}. Beyond routing optimization, healthcare logistics requires integrated frameworks that support personnel allocation, automated sensing for inventory management, and adaptive supply chain control capable of responding to changing patient needs and clinical urgency. While prior work in drone logistics has explored last-mile planning, facility siting, and fleet coordination \cite{garg2023lastmile,sushma2025spatial,hu2024rloverview,ning2024marlsurvey}, a key research gap remains in developing learning-based systems that jointly address clinical priority, strict delivery deadlines, payload constraints, and stochastic task arrivals under limited communication and information availability.

Traditional optimization methods such as mixed-integer programming, metaheuristics, and genetic algorithms are effective for static UAV routing but often fail to adapt efficiently to dynamic medical supply requests with heterogeneous urgency levels. Each new task typically requires costly re-optimization, limiting their scalability for real-time healthcare logistics \cite{garg2023lastmile,sushma2025spatial}. Prior multi-UAV systems, including our own previous work \cite{guven2024adhoc}, leverage evolutionary approaches effectively in fixed-task settings but suffer from computational inefficiency when applied to highly dynamic, time-sensitive environments.

Recent advances in UAV routing and multi-agent reinforcement learning (MARL) demonstrate strong potential for scalable, adaptive decision-making. Wang \textit{et al.} \cite{wang_c-sppo_2025} introduced the C-SPPO framework, a centralized reinforcement learning model that minimizes flight conflicts and delivery delays in large-scale logistics routing. Cui \textit{et al.} \cite{cui_design_2025} proposed a GCN-based policy network that improves multi-UAV task allocation efficiency under distance constraints, while Qiu \textit{et al.} \cite{qiu_distributed_2025} developed a distributed cooperative UAV search and rescue framework robust to communication limitations. Gabler and Wollherr \cite{gabler_decentralized_2024} emphasized decentralized actor–critic structures to enhance scalability and real-world deployability, and Kong and Sousa \cite{kong_piggybacking_2024} demonstrated how UAVs can simultaneously perform package delivery and wireless coverage through deep Q-learning-based trajectory control. Recent surveys \cite{ning2024marlsurvey} have highlighted ongoing challenges in integrating autonomy, coordination, and real-world uncertainty. 


MARL offers significant potential for adaptive decision-making in such contexts, yet designing a general framework for heterogeneous tasks that are dynamically assigned has not been investigated yet. This paper addresses the gap between existing MARL approaches and practical medical supply applications by presenting a unified learning-based framework for adaptive healthcare logistics optimization. We introduce a multi-agent reinforcement learning model that operates under stochastic demand, partial fleet observability, and strict delivery deadlines for real-world medical UAV operations.

To examine the trade-off between throughput and coordination performance, we also evaluated two distributed actor–learner architectures—Asynchronous PPO (APPO) \cite{asynchPPO} and IMPALA \cite{espeholt2018impalascalabledistributeddeeprl} and a classical actor-critic method (A2C) \cite{a2c}. These frameworks are designed for large-scale environments where many agents collect experience in parallel. 

The contributions of this paper are as follows.
\begin{itemize}
\item A partially observable MDP formulation for multi-UAV medical delivery with full task visibility but partial fleet position awareness,  modeling depot resupply, stochastic task arrivals, and clinical urgency.

\item A reward shaping framework with proximity guidance, distance reduction bonuses, and urgency-based weighting that accelerates learning with minimal computational overload.

\item Experimental analysis with various MARL methods for observing the effect of network architecture, policy-update mechanism, and data collection in dynamic delivery missions.
\end{itemize}

The paper is organized as follows. Section \ref{sec:system} presents the problem formulation. Section \ref{sec:marl} details the MARL framework with observation design and reward structure. Section \ref{sec:results} presents experimental results. Section \ref{sec:conc} concludes.

\section{System Model}
\label{sec:system}

This study models the coordination of multiple UAVs for real-time delivery of medical supplies in an urban environment. The system includes the operational characteristics of healthcare logistics: Dynamic demand, time-critical deliveries, and limited UAV resources. The framework provides the mathematical model for the reinforcement learning formulation presented in Section~\ref{sec:marl}.

\subsection{Environment Representation}

Table~\ref{tab:params} summarizes the environment parameters. The environment is represented as a grid-based graph \( G = (V, E) \), where each cell corresponds to a vertex \( v \in V \), and each UAV can move to one of its four neighboring cells at each time step.

The main components of the system are:
\begin{itemize}
    \item Depots \( D \subseteq V \): nodes where UAVs collect supplies and refuel.
    \item Clinics \( H \subseteq V \): nodes where delivery requests originate.
    \item UAV fleet \( U = \{1, 2, \dots, N\} \): each UAV has a maximum payload capacity \( P_{max} \) and a discrete position \( x_i(t) \in V \).
    \item Delivery tasks \( \mathcal{T}(t) \): dynamically appearing requests requiring pickup from a depot and delivery to a clinic before a deadline.
\end{itemize}

\begin{table}[t!]
\centering
\scriptsize
\caption{Summary of Environment Parameters}
\begin{tabular}{lll}
\toprule
\textbf{Symbol} & \textbf{Description} & \textbf{Value} \\
\midrule
$N$ & Number of UAVs & 5–20 \\
$G$ & Grid dimensions & $30 \times 30$ \\
$c$ & Cell size (m) & $400$ m \\
$v$ & UAV speed (m/s) & $50$ m/s \\
$P_{\max}$ & Max payload & $5$ units \\
$R_{\text{comm}}$ & Comm. range & $400$ m \\
$T_{\max}$ & Max episode steps & $200$ \\
$\lambda$ & Task arrival rate & $0.1$–$0.3$ \\
$K_{\max}$ & Max active tasks & $10$ \\
$p(t)$ & UAV payload level & $0$–$P_{\max}$ \\
$x_i(t)$ & UAV position & Grid cell \\
$\tau$ & Delivery task & — \\
$u$ & Urgency class & \{crit., urg., std.\} \\
$d_{\text{crit}}$ & Critical deadline & $10$ steps \\
$d_{\text{urg}}$ & Urgent deadline & $20$ steps \\
$d_{\text{std}}$ & Standard deadline & $50$ steps \\
$H$ & Number of hospitals & $4$ \\
$D$ & Number of depots & $2$ \\
$I_0$ & Initial inventory & $10$ units \\
$T_d$ & Pickup/delivery time & $5$ s \\
$\rho$ & Consumption rate & $0.1$ \\
\bottomrule
\end{tabular}
\label{tab:params}
\end{table}

\begin{figure}[b!]
    \centering
    \includegraphics[width=0.99 \linewidth]{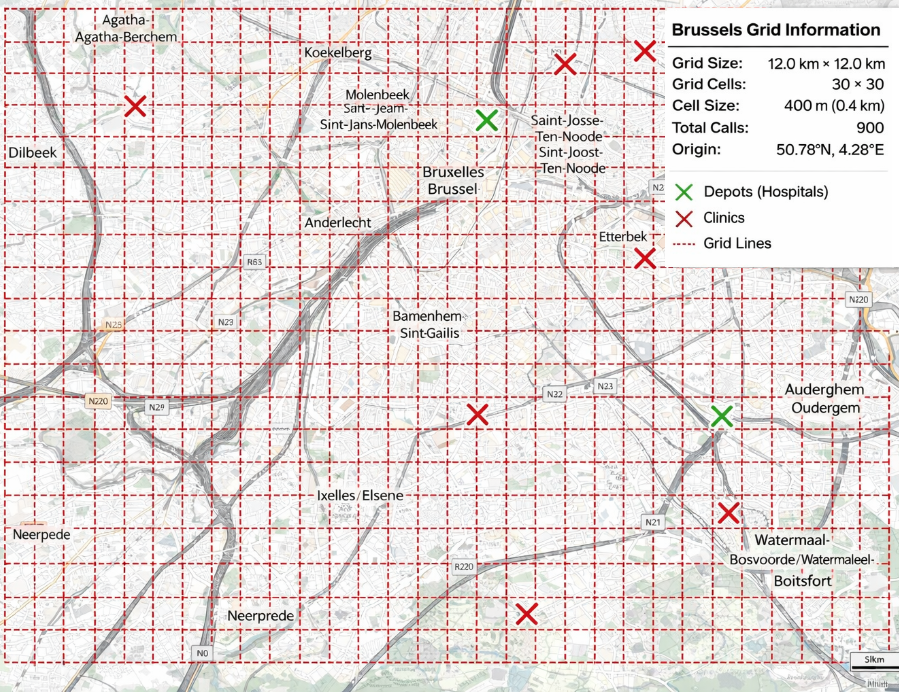}
    \caption{Operational region centered around the Brussels Capital Region. The grid defines UAV navigation cells with depots (green) and clinics (red).}
    \label{fig:region_map}
\end{figure}

\subsection{Case Study: Brussels Operational Region}

Figure~\ref{fig:region_map} illustrates the case study region centered on the Brussels Capital Area. The area is modeled as a $12\,\text{km} \times 12\,\text{km}$ grid, divided into $30 \times 30$ cells, each representing a 400\,m$\times$400\,m area. Green markers indicate depots (hospitals with storage and refilling capacity), while red markers represent clinics that request supplies.

\subsection{Medical Dynamic Task Model}
\label{subsec:medical_task_model}

Each delivery request is represented as a medical task $\tau$ defined by a pickup location $p_{\text{source}}$, a target hospital $p_{\text{target}}$, an urgency class $u$, a creation time $t_{\text{created}}$, and a deadline $t_{\text{deadline}}$:
\[
\tau = (p_{\text{source}}, p_{\text{target}}, u, t_{\text{created}}, t_{\text{deadline}}).
\]

New tasks arrive stochastically with probability $\lambda$ at each time step, reflecting the irregular and unpredictable nature of clinical demand. The urgency level $u \in \{\text{critical}, \text{urgent}, \text{standard}\}$ determines the feasable delivery window. The deadline is assigned as
\[
t_{\text{deadline}} = t_{\text{created}} + \Delta(u),
\]
where $\Delta(u)$ is the allowed time for the task and $\Delta_{\text{crit}} < \Delta_{\text{urg}} < \Delta_{\text{std}}$. Critical tasks correspond to life-saving items such as blood units or emergency medication, whereas standard tasks represent routine consumables with more flexible timing.

Hospitals maintain initial inventories $I_0$ that decrease at each step according to a consumption rate $\rho$, which models continuous clinical use of supplies:
\[
I_h(t+1) = \max\Big(0,\; I_h(t) - \rho\big(1 + \tfrac{|P_h(t)|}{10}\big)\Big),
\]
where $P_h(t)$ denotes the set of patients waiting at hospital $h$. As inventories decline, hospitals generate new tasks with urgency linked to patient conditions and required treatment types. Patients enter a waiting list upon arrival, each inheriting a personal deadline $d_u$. If treatment cannot begin before this deadline, the patient is considered deceased, and a mortality penalty is applied to the learning agent.

Each task requires UAVs to pick up a supply package at a depot and deliver it to the designated hospital. Depots broadcast available tasks to all UAVs, agents are tasked with choosing actions based on urgency, distance, and remaining time to deadline.

\subsection{UAV Operations and Constraints}

Each UAV follows a periodic process:
\begin{enumerate}
    \item Travel to a depot and collect available supplies.
    \item Pick up an assigned delivery task if available.
    \item Transport the package to the corresponding clinic.
    \item Refill payload at a depot if capacity is low.
\end{enumerate}

The UAV’s state at time \( t \) includes its grid position \( x_i(t) \), remaining payload \( p_i(t) \), and any assigned task \( \tau_i(t) \).  Movements are limited to one adjacent cell per step.

Payload evolves according to:
\[
p_i(t+1) =
\begin{cases}
P_{max}, & \text{if refilling at depot}, \\
p_i(t) - 1, & \text{if picking up a task}, \\
p_i(t), & \text{otherwise.}
\end{cases}
\]

A UAV can carry at most one active task at a time and must deliver it before its deadline. UAVs fly on constant altitude and each delivery or pick-up is accounted for extra flight time $T_d$ due to altitude changes. 

Communications assume a disc model. When two UAVs are within a distance $R_{comm}$, they can communicate and share their information to each other. For battery considerations, a flight time limit is given from take-off until landing. For reference, we assumed the batteries have a capacity of 0.5kWh and each movement and delivery costs 0.8Wh energy for our test scenario. A more detailed analysis on battery and altitude considerations will be part of future work.

\subsection{Objective Function}

The system aims to maximize the overall delivery performance across all UAVs and tasks. The total objective balances three key components:

\[
J = R_{\text{deliveries}} + R_{\text{urgency}} - C_{\text{delays}} - C_{\text{inefficiency}}.
\]

Here:
\begin{itemize}
    \item \( R_{\text{deliveries}} \): rewards for successful and timely deliveries.
    \item \( R_{\text{urgency}} \): additional bonuses for critical and urgent tasks.
    \item \( C_{\text{delays}} \): penalties for late or expired deliveries.
    \item \( C_{\text{inefficiency}} \): penalties for unnecessary movement or idling.
\end{itemize}

This formulation encourages UAVs to complete high-priority deliveries first while maintaining efficiency in motion and resource usage.

At each time step, the system must decide how UAVs should move and which task to pick up. These sequential and uncertain decisions form a dynamic control problem.  To solve this, the UAV coordination process is represented as a Markov decision process (MDP), where the state captures UAV positions, payload levels, and current task information. Section~\ref{sec:marl} reformulates this system model within the reinforcement learning framework, defining the corresponding states, actions, transition dynamics, and reward structure used for multi-agent training.

\section{Reinforcement Learning Formulation}
\label{sec:marl}

\subsection{Markov Decision Process Specification}

The delivery task is formulated as a partially observable Markov decision process:
\[
\mathcal{M} = \langle \mathcal{N}, \mathcal{S}, \{\mathcal{A}_i\}, P, R, \{\Omega_i\}, O, \gamma \rangle,
\]
where:
\begin{itemize}
\item $\mathcal{N}$ is the set of UAVs,
\item $\mathcal{S}$ is the global state (all UAV positions, all task states, time),
\item $\mathcal{A}_i = \{\text{up}, \text{down}, \text{left}, \text{right}, \text{stay}\}$ is agent $i$'s action space,
\item $R: \mathcal{S} \times \mathcal{A}^N \to \mathbb{R}$ is the reward function,
\item $\Omega_i$ is agent $i$'s observation space,
\item $O: \mathcal{S} \times i \to \Omega_i$ is the observation function,
\item $\gamma = 0.99$ is the discount factor.
\end{itemize}

\subsection{Observation Space}

Each UAV agent receives a compact observation vector at each time step. This vector provides the information required to make navigation and task allocation decisions while coordinating with other UAVs in the environment.

The observation includes six main components:

\begin{itemize}
    \item \textit{Positional information:} Normalized positions of UAVs based on last communication time. 
    \item \textit{Own state:} Information about UAV’s internal status, including its current payload level (normalized between 0 and 1) and whether it is  carrying a delivery task.
    
    \item \textit{Closest pending task:} Information about the nearest unassigned delivery request, such as the relative position of the pickup depot and destination hospital, the task’s urgency level, and the remaining time before its deadline. This helps the UAV decide which pending task to prioritize based on distance and urgency.
    
    \item \textit{Current carried task:} Details about the task currently being delivered, including the relative position of the delivery target and the time remaining until the deadline. If the UAV is not carrying any task, this part of the observation remains zero.
    
    \item \textit{Closest depot:} The relative position of the nearest depot, allowing the UAV to plan refilling or resupply actions when its payload is low.
    
    \item \textit{Closest hospital:} The relative position of the nearest hospital or clinic, which supports decisions related to future deliveries or route planning.
    
    \item \textit{Global context:} General information about the environment, including the total number of active tasks, the proportion of pending tasks, and the normalized simulation time. This provides situational awareness and helps balance workload distribution across the UAV fleet.
\end{itemize}

Together, these components allow each agent to maintain awareness of its own operational status, nearby opportunities for delivery, and the overall mission context.

\begin{table}[t]
\centering
\scriptsize
\caption{Reward structure for UAV medical delivery tasks.}
\label{tab:rewards}
\begin{tabular}{p{2.2cm} p{3.2cm} r}
\toprule
\textbf{Category} & \textbf{Description} & \textbf{Value} \\
\midrule
\multicolumn{3}{l}{\textit{Clinical outcomes (sparse rewards)}} \\
\midrule
Delivery completion & Successful delivery & +50.0 \\
Critical delivery bonus & Critical task completed & +20.0 \\
Urgent delivery bonus & Urgent task completed & +10.0 \\
Early delivery bonus & Remaining time before deadline & +5.0 × ratio \\
Deadline violation & Late or missed delivery & –15.0 \\
\midrule
\multicolumn{3}{l}{\textit{Task discovery and progress (dense rewards)}} \\
\midrule
Task proximity & Near pending task & +0.2 × proximity \\
Pickup success & Task picked up at depot & +5.0 \\
Urgent task claim & Claim of urgent/critical task & +3.0 \\
Distance reduction & Moved closer to target & +0.3 × distance gain \\
Progress movement & Step toward assigned target & +0.5 \\
\midrule
\multicolumn{3}{l}{\textit{Resource management and penalties}} \\
\midrule
Refill at depot & Refill when payload is low & +1.0 \\
Depot visit (low) & Visit depot when half-empty & +2.0 \\
Movement cost & Per movement step & –0.001 \\
Idle penalty & Idle away from depot & –0.01 \\
Mortality penalty & Expired critical task & –20.0 \\
\bottomrule
\end{tabular}
\end{table}

\subsection{Reward Shaping for Medical Delivery}

The learning process combines sparse clinical rewards with dense shaping rewards to guide UAV agents toward efficient and meaningful behaviors. Table~\ref{tab:rewards} summarizes the reward components used during training. This reward design encourages UAV agents to prioritize critical medical deliveries while maintaining efficient resource use and coordinated movement. The largest rewards are given upon successful and timely deliveries. Urgency-based bonuses are also given to prioritize critical and urgent tasks.  

Dense shaping rewards provide continuous feedback even when deliveries are not yet completed. They help agents learn useful intermediate behaviors such as moving toward pending tasks, reducing distance to delivery targets, and refilling supplies before depletion. Small penalties for unnecessary movement and idling discourage inefficient actions. Finally, the mortality penalty models the severe consequence of missing critical deliveries, reinforcing the importance of meeting medical deadlines within the learning process. 

\subsection{Action Space}

Each UAV executes one of five discrete actions per timestep:
\[
\mathcal{A}_i = \{\text{up}, \text{down}, \text{left}, \text{right}, \text{stay}\}.
\]
Actions move the UAV one grid cell or keep it stationary. The environment handles task claiming, pickup, delivery, and refilling automatically when position and state conditions are satisfied. We used a discrete action space instead of a continuous model in order to focus on the long-term planning aspect of our model. Furthermore, with cell movements being synchronized, UAVs only exchange information when all UAVs reach their waypoints, which decreases communication complexity and allows for synchronized updates.

\subsection{Training Algorithms}

We implemented a family of policy gradient algorithms using Ray RLlib \cite{liang2018rllibabstractionsdistributedreinforcement} to study both architectural and systems-level design choices. The synchronous on-policy baseline is Proximal Policy Optimization (PPO), instantiated with a three-layer multilayer perceptron (MLP) policy. Two architectural variants of PPO were considered:
\begin{itemize}
    \item \textbf{PPO Large FCNet}: a deeper fully connected network with hidden layers of size $[512, 512, 256]$ sharing parameters between actor and critic. This variant tests whether additional capacity improves coordination under the same on-policy update rule.
    \item \textbf{PPO LSTM}: a recurrent policy based on an LSTM-containing RLModule with stacked dense layers and a 256-unit LSTM cell. This configuration models temporal dependencies such as task queues and approaching deadlines.
\end{itemize}

To provide a classical on-policy baseline, we also include Advantage Actor–Critic (A2C), representing a low-complexity alternative to PPO. Finally, we evaluate two distributed actor–learner architectures:

\begin{itemize}
    \item \textbf{APPO}: Asynchronous PPO with V-trace corrections and centralized learners. Experience is collected in parallel from multiple workers and consumed off-policy.
    \item \textbf{IMPALA}: an importance-weighted actor–learner architecture optimized for high-throughput sampling with V-trace targets.
\end{itemize}

All methods share the same observation and action spaces, reward structure (Table~\ref{tab:rewards}), and discount factor $\gamma = 0.99$. Hyperparameters for each algorithm (batch sizes, entropy regularization, clipping coefficients, and gradient clipping) use RLlib defaults. Multi-agent training uses a centralized policy mapping in which each UAV runs its own copy of the policy network which is evaluated on a decentralized setting.

\section{Results and Discussion}
\label{sec:results}

\subsection{Experimental Setup}

The experimental evaluation was conducted using a 30 by 30 grid representing a 12 km by 12 km urban area centered on Brussels, with each cell spanning 400 meters. The infrastructure consisted of 2 supply depots and 6 clinic locations. UAV parameters include maximum payload of 5 units, movement speed of 50 meters per second, and communication range of 400 meters. Medical supply requests arrived stochastically with rate 0.2 per timestep, categorized into three urgency levels: critical (15\%, $\Delta_{\text{crit}}=5$ steps), urgent (35\%, $\Delta_{\text{urg}}=10$ steps), and standard (50\%, $\Delta_{\text{std}}=20$ steps). Each UAV has 200 steps of movement until they have to return to their initial positions. Each mission ends when at least 15 tasks are completed and all tasks that have been assigned must be completed. Since the arrival of tasks are nondeterministic, some missions can have significantly more tasks and more urgency, which represents the real-life environment in terms of a health crisis.

We use Ray RLlib \cite{liang2018rllibabstractionsdistributedreinforcement} with four algorithms (A2C, PPO, APPO, IMPALA) using 3-layer MLP architecture (512-256-128 units), Adam optimizer with learning rate 0.0003, and discount factor 0.99. Each configuration trained for 2,000,000 steps using 8 parallel workers with 2 environments each on our 32 core Intel Xeon Gold 6444Y CPU. Evaluation consists of 1000 episodes per configuration with action selection across fleet sizes of 4, 8, 12, and 16 UAVs. 

\subsection{Computational Analysis}

\begin{figure}[t!]
    \centering
    \includegraphics[width=1\linewidth]{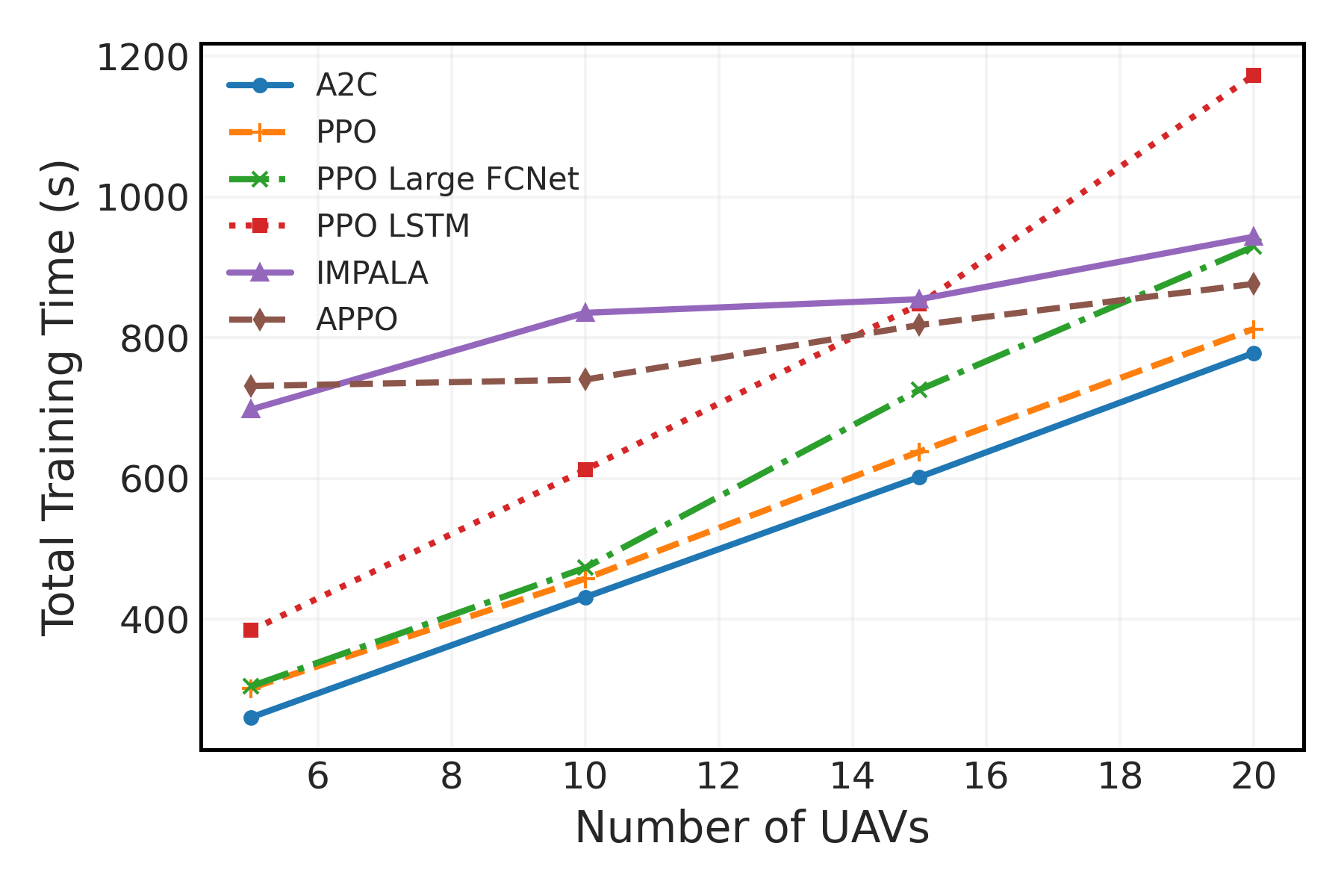}
    \caption{Training Times for each algorithm}
    \label{fig:training_times}
\end{figure}

Fig. \ref{fig:training_times} shows training times in seconds for each algorithm. Asynchronous models are trained around 900 seconds regardless of number of agents whereas classical models have increasing training times from 350 to 1200 seconds. 

Evaluation times for a single episode range from 0.5 to 1.2 seconds depending on episode length for all algorithms. LSTM-based PPO models take 3 seconds per episode due to more complex model structure.

Fast training times allow the existing model to be quickly trained on new scenarios whereas fast evaluation requires minimal computational capacity that can be used on most UAV processors.

\begin{figure}[b!]
    \centering
    \includegraphics[width=1\linewidth]{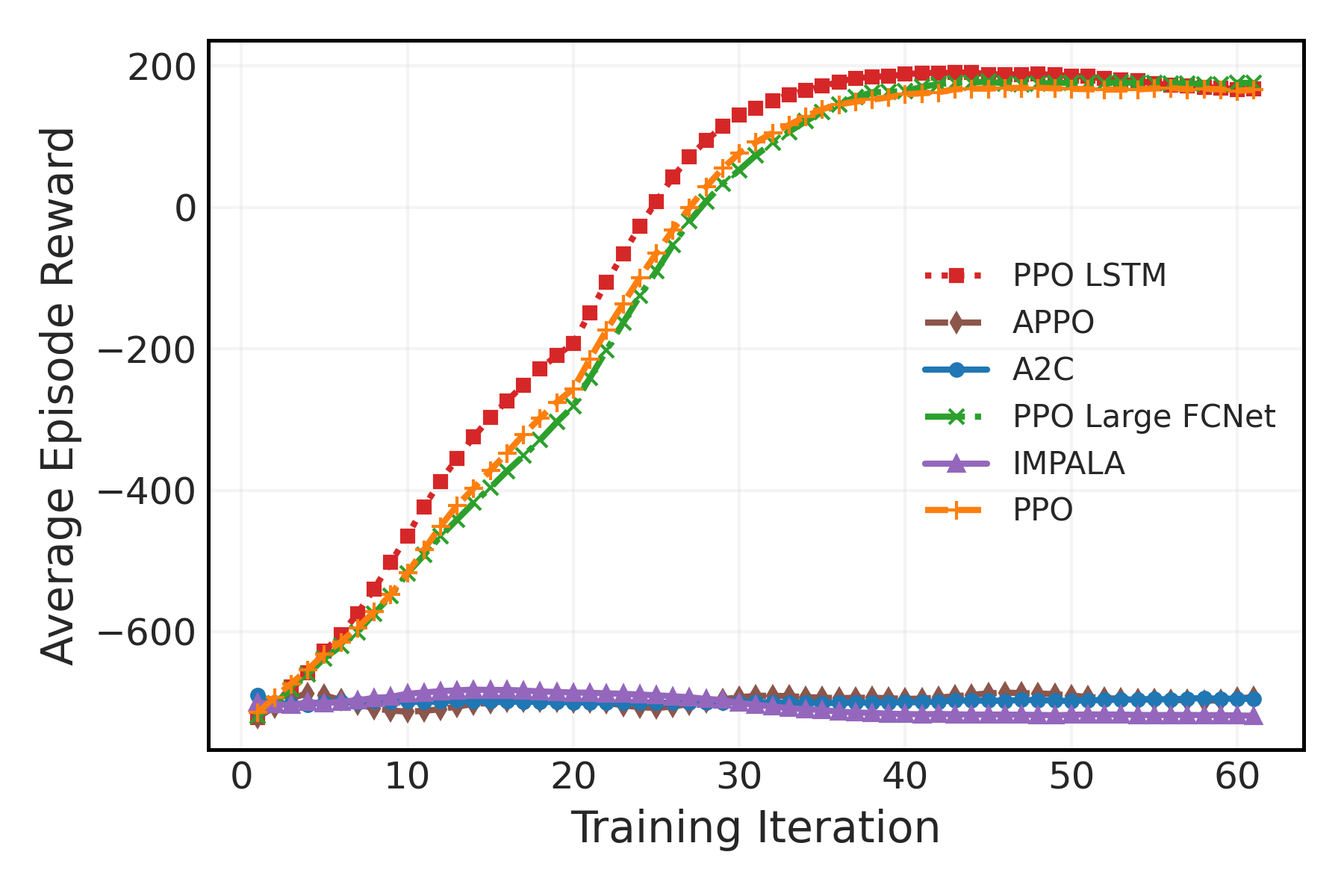}
    \caption{Episode return mean.}
    \label{fig:episode_return_mean}
\end{figure}

\begin{figure*}[t!]
    \centering
    \hfill
    \begin{subfigure}[b]{0.49\textwidth}
        \centering
        \includegraphics[width=1\textwidth]{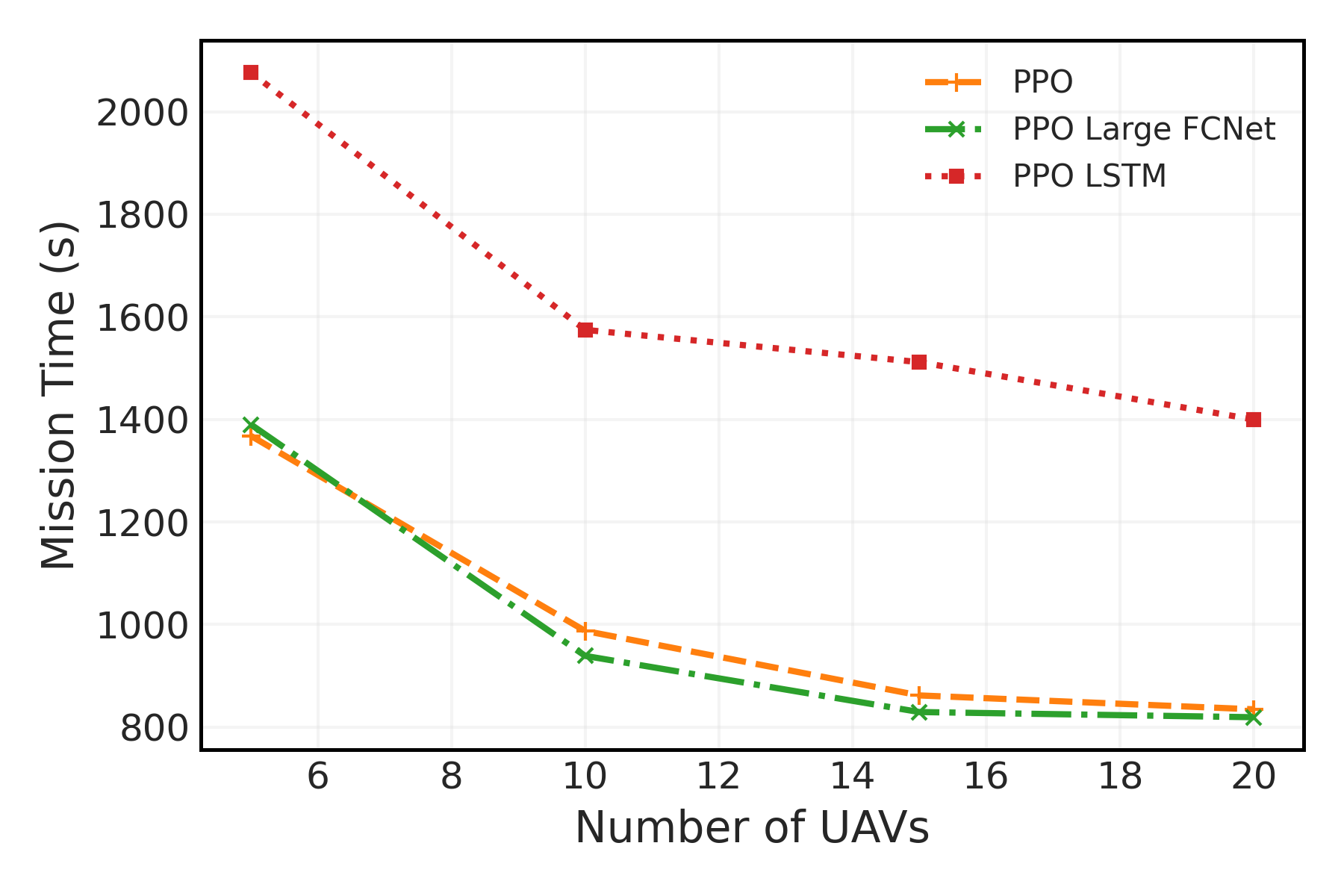}
        \caption{}
        \label{fig:time_comp}
    \end{subfigure}
    \hfill
    \begin{subfigure}[b]{0.49\textwidth}
        \centering
        \includegraphics[width=1.0\textwidth]{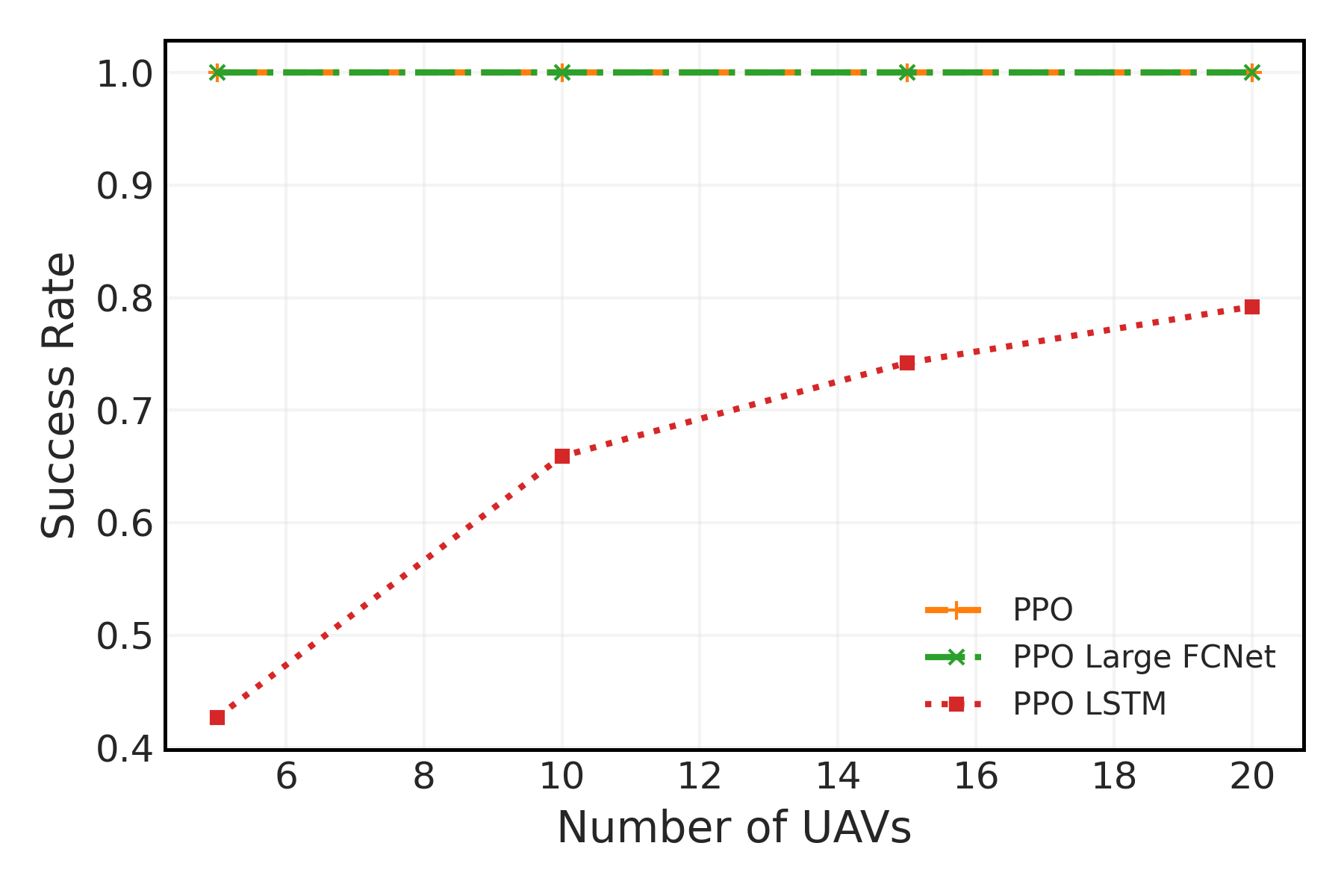}
        \caption{}
        \label{fig:succ_comp}
    \end{subfigure}
    \caption{Performance comparison across different configurations of PPO for varying fleet sizes. (a) Average mission time across UAV counts. (b) Success rate for different RL algorithms.}
    \label{fig:model_comparison}
\end{figure*}

\subsection{Learning Performance}

We first examine learning dynamics for a fixed fleet size. Fig.~3 shows the evolution of episode returns for 10 UAVs over 2,000,000 training steps. PPO exhibits clear convergence, improving from an initial average return of approximately $-600$ to around $-200$ as training progresses. In contrast, APPO and IMPALA remain close to their initial performance and fail to achieve meaningful learning progress in this domain. This behavior indicates that in our setting, simple actor--critic updates (A2C) and off-policy V-trace corrections (APPO, IMPALA) are insufficient to stabilize learning under the combined challenges of strict delivery deadlines, stochastic task arrivals, and cooperative multi-agent assignments. 

The clipped policy updates in PPO, together with carefully shaped rewards that penalize missed deadlines and inefficient routing, enable more stable policy improvement and encourage exploration of coordinated delivery strategies. As training progresses, the learned policies gradually reduce mission completion time while increasing the fraction of successfully delivered medical supplies. This behavior suggests that PPO effectively balances exploration and exploitation in environments where agents must jointly allocate limited resources under time constraints.

From a systems perspective, these results highlight the importance of stable on-policy learning mechanisms for coordinating UAV fleets in time-critical healthcare logistics environments. The findings further indicate that policy stability plays a crucial role in multi-agent scheduling problems where delivery deadlines, spatial constraints, and agent cooperation must be handled simultaneously. Consequently, the proposed MARL framework provides a promising approach for enabling reliable and scalable UAV-assisted medical logistics in dynamic operational settings.

\subsection{Mission Performance}

Figure~\ref{fig:model_comparison} shows how algorithms scale with fleet size. PPO success rate is $100\%$ for all fleet sizes with mission time decreasing from 1400~s to 800~s. We observe that effective workload distribution and coordination benefit from additional agents, as reflected in the decline of average mission times. Furthermore, we can observe that 15 UAVs provide nearly identical results to 20 UAVs. Large architecture variant of PPO (PPO Large FCNet) follows the similar trend and closely track standard PPO in terms of mission-level performance with minor improvements. LSTM provided poorer results whereas using a larger fully connected network on standard PPO improved the performance. This indicates that the benefits from sequential actions are minimal and the mission requires more adaptive decision-making.

\section{Conclusion}
\label{sec:conc}

This paper presented a MARL framework for time-critical UAV medical supply delivery in urban environments with stochastic task arrivals and heterogeneous urgency levels. Using a realistic grid representation of a $12\,\text{km} \times 12\,\text{km}$ city environment with multiple depots and clinics, we evaluated several reinforcement learning algorithms for coordinating UAV fleets under payload, communication, and deadline constraints.

Experimental results demonstrate that synchronous on-policy learning using PPO consistently achieves reliable coordination across different fleet sizes. In particular, PPO converges from an initial average return of approximately $-600$ to around $-200$ during training and achieves a $100\%$ task completion rate during evaluation. Mission duration decreases significantly as fleet size increases, dropping from approximately $1400\,\text{s}$ with smaller fleets to about $800\,\text{s}$ with larger fleets due to improved workload distribution among agents. In contrast, asynchronous approaches such as APPO and IMPALA fail to achieve meaningful convergence in this environment, highlighting the importance of stable on-policy updates and reward shaping for cooperative UAV coordination under strict deadlines.

Computational analysis further shows that training can be completed within practical time scales, with asynchronous models requiring roughly $900\,\text{s}$ and classical models ranging between $350\,\text{s}$ and $1200\,\text{s}$ depending on fleet size. Evaluation time remains low ($0.5$--$1.2\,\text{s}$ per episode), indicating that the learned policies can be executed on resource-constrained UAV platforms in real time. These findings demonstrate that PPO-based coordination provides a practical and scalable approach for UAV-assisted medical logistics during emergencies.

\section*{Acknowledgment}
This work was supported by the Brains for Brussels research and innovation funding program of the Région de Bruxelles-Capitale–Innoviris under Grant RBC/BFB 2023-BFB-2.

\bibliographystyle{IEEEtran}


\end{document}